\documentclass[11pt]{article}

\usepackage[a4paper,margin=1in]{geometry}
\usepackage[utf8]{inputenc}
\usepackage[T1]{fontenc}
\usepackage{setspace}
\usepackage{parskip}
\usepackage{graphicx}
\usepackage{xcolor}
\usepackage{titlesec}
\usepackage{booktabs}
\usepackage{multirow}
\usepackage{fancyhdr}
\usepackage{natbib}
\usepackage{amsmath,amsfonts,amsthm,amssymb}
\usepackage{hyperref}
\usepackage{cleveref}
\usepackage{lmodern}
\usepackage{tcolorbox}
\usepackage{csquotes}

\definecolor{sfblue}{HTML}{00A1E0}      
\definecolor{sfnavy}{HTML}{032D60}      
\definecolor{sfgray}{HTML}{706E6B}      
\definecolor{sflightblue}{HTML}{EAF5FC} 

\hypersetup{
  colorlinks=true,
  linkcolor=sfblue,
  citecolor=sfnavy,
  urlcolor=sfblue
}

\renewcommand{\headrulewidth}{0.4pt} 
\renewcommand{\footrulewidth}{0pt} 

\fancypagestyle{plain}{
  \fancyhf{} 
  \fancyfoot[C]{\textcolor{sfgray}{\thepage}} 
  \renewcommand{\headrulewidth}{0pt}
}

\usepackage{xspace}

\newcommand{\github}{\raisebox{-1.5pt}{\includegraphics[height=1.05em]{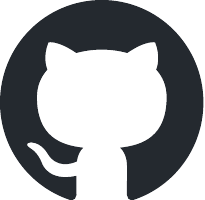}}\xspace}

\newcommand{\huggingface}{\raisebox{-1.5pt}{\includegraphics[height=1.05em]{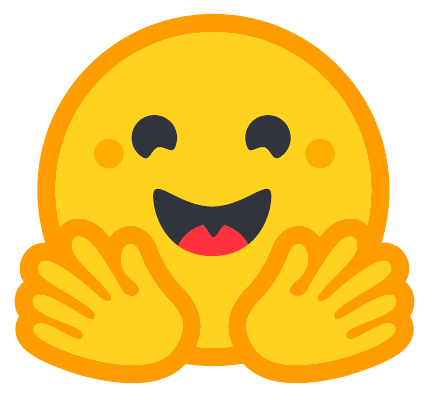}}\xspace}

\makeatletter
\renewcommand{\maketitle}{
  \thispagestyle{plain} 
  \noindent
  \vspace*{-10pt}
  \noindent\raisebox{0pt}[0pt][0pt]{\includegraphics[height=1cm]{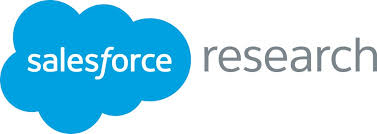}}

  \vspace{-5pt}
  \color{sfgray}\rule{\linewidth}{0.6pt}

  \vspace{10pt}
  {\Huge\bfseries\color{sfnavy} \@title \par}
  \vspace{0.5em}
  {\large \@author \par}
  \vspace{0.2em}
}
\makeatother

\titleformat{\section}{\large\bfseries\color{sfnavy}}{\thesection}{1em}{}
\titleformat{\subsection}{\normalsize\bfseries\color{sfnavy}}{\thesubsection}{1em}{}

\title{\huge CoDA: Coding LM via Diffusion Adaptation}
\date{\today}

\usepackage{authblk}

\makeatletter
\renewcommand\AB@affilsepx{ \quad \protect\Affilfont}
\makeatother

\author[$\ast,\dagger$]{Haolin Chen}
\author[$\ast,\dagger$]{Shiyu Wang}
\author[$\ast,\dagger$]{Can Qin}
\author[$\dagger$]{Bo Pang}
\author[$\dagger$]{Zuxin Liu}
\author[$\dagger$]{Jielin Qiu}
\author[$\dagger$]{Jianguo Zhang}
\author[$\dagger$]{Yingbo Zhou}
\author[$\dagger$]{Zeyuan Chen}
\author[$\dagger$]{Ran Xu}
\author[$\dagger$]{Shelby Heinecke}
\author[$\dagger$]{Silvio Savarese}
\author[$\dagger$]{Caiming Xiong}
\author[$\dagger$]{Huan Wang}
\author[$\dagger$]{Weiran Yao}

\affil[$\dagger$]{Salesforce AI Research}
\affil[$\ast$]{Core Contributors}

\usepackage{amsmath,amsfonts,bm}

\def\eqref#1{equation~\ref{#1}}

\def\1{\bm{1}}

\def\vmu{{\bm{\mu}}}

\def\ve{{\bm{e}}}

\def\vx{{\bm{x}}}

\def\mI{{\bm{I}}}

\def\mQ{{\bm{Q}}}
\def\mR{{\bm{R}}}

\def\mSigma{{\bm{\Sigma}}}

\DeclareMathAlphabet{\mathsfit}{\encodingdefault}{\sfdefault}{m}{sl}
\SetMathAlphabet{\mathsfit}{bold}{\encodingdefault}{\sfdefault}{bx}{n}

\newcommand{\E}{\mathbb{E}}

\begin{document}

\maketitle

\begin{tcolorbox}[colback=sflightblue!60,
                  colframe=sfblue,
                  boxrule=0.6pt,
                  arc=2mm,
                  left=6pt,right=6pt,top=6pt,bottom=6pt]
\textbf{Abstract.} Diffusion language models promise bidirectional context and infilling capabilities that autoregressive coders lack, yet practical systems remain heavyweight. We introduce CoDA, a 1.7B-parameter diffusion coder trained on TPU with a fully open-source training pipeline. CoDA pairs large-scale diffusion pre-training with code-centric mid-training and instruction tuning, enabling confidence-guided sampling that keeps inference latency competitive. On Humaneval, MBPP, and EvalPlus, CoDA-1.7B-Instruct matches or surpasses diffusion models up to 7B parameters. Our release includes model checkpoints, evaluation harnesses, and TPU training pipelines to accelerate research on lightweight diffusion-based coding assistants.
\medskip
\begin{center}
\begin{tabular}{rll}
    \github & \textbf{\small{Code}:} & 
    \href{https://github.com/SalesforceAIResearch/CoDA}{\small{\texttt{https://github.com/SalesforceAIResearch/CoDA}}}\\
    \huggingface & \textbf{\small{Model}:} & \href{https://huggingface.co/Salesforce/CoDA-v0-Instruct}{\small{\texttt{https://huggingface.co/Salesforce/CoDA-v0-Instruct}}}
\end{tabular}
\end{center}
\end{tcolorbox}

\color{black}

\section{Introduction}

Large language models (LLMs) have rapidly advanced automatic code generation, powering developer assistants that translate natural language intent into working programs. Open-source models such as StarCoder~\citep{li2023starcoder} and Qwen3-Coder~\citep{yang2025qwen3} exemplify the progress of the autoregressive (AR) paradigm, which predicts code token-by-token. Despite their success, AR decoders propagate errors sequentially, struggle to leverage context bidirectionally, and falter on tasks such as filling in missing code segments or editing large spans of text.

Diffusion language models (DLMs) offer an attractive alternative~\citep{nie2025large, gong2025diffucoder, ye2025dream}. Instead of generating each token sequentially, DLMs generate sequences through an iterative denoising process: a forward stage corrupts text by adding noise (e.g., masking tokens), and a backward stage gradually reconstructs the original sequence. This two-phase approach enables parallel token generation and bidirectional context awareness—i.e. the model can utilize both left and right context when generating each part of the code—addressing limitations of purely left-to-right generation. Diffusion LLMs also offer enhanced controllability and naturally support text infilling, a desirable feature for code completion and editing tasks~\citep{gong2025diffucoder}. For instance, a diffusion model can fill in a missing block in the middle of a code file by considering the surrounding context, which poses a challenge for standard AR models.

Prior work has validated the feasibility of scaling diffusion generation to natural language and code, yet current DLM coders typically target large model budgets (e.g., 7B--8B parameters). We introduce \textbf{CoDA}, a 1.7B-parameter diffusion language model for code that is trained on an efficient TPU pipeline. CoDA adapts the Qwen3 backbone~\citep{yang2025qwen3} to a diffusion objective, combines $\sim$180B tokens of general pre-training with $\sim$20B tokens of curated code, and applies supervised fine-tuning for instruction following. We release the model weights, evaluation harness, and training recipes, enabling the community to reproduce diffusion-based coding models without proprietary tooling.

Our main contributions are as follows:
\begin{enumerate}
    \item We develop a fast 1.7B diffusion coder, demonstrating that compact DLMs can deliver interactive latency while retaining the benefits of bidirectional decoding.
    \item We show that CoDA achieves competitive pass@1 scores relative to diffusion and autoregressive models up to 7B parameters, narrowing the performance gap while using one quarter of the weights.
    \item We open-source a high-performance TPU training pipeline that lowers the barrier to scaling future DLMs.
\end{enumerate}

\section{Related Works}
AR code models remain the default choice for open-source assistants~\citep{li2023starcoder,yang2025qwen3}, yet they decode tokens sequentially, one at a time. DLMs tackle these limitations by iteratively refining noisy token sequences, inheriting ideas from continuous generative diffusion while adapting them to discrete vocabularies~\citep{lou2023discrete,shi2024simplified}. The resulting masked-diffusion objectives and ratio-matching estimators provide the methodological footing needed for scaling text diffusion.

Two complementary development tracks have emerged for large DLMs. Training bespoke diffusion coders from scratch yields systems such as SMDM~\citep{nie2024scaling} and the LLaDA family~\citep{nie2025large}. In parallel, adaptation techniques repurpose pretrained AR checkpoints into diffusion decoders—DiffuGPT and DiffuLLaMA illustrate this recipe~\citep{gong2024scaling}—enabling rapid bootstrapping of strong models like Dream~\citep{ye2025dream}. DiffuCoder extends the approach with GRPO-based reinforcement learning~\citep{gong2025diffucoder}, while parallel works~\citep{zhao2025d1,wang2025revolutionizing} further advance the use of reinforcement learning in DLMs.

Inference optimization has followed suit. KV-cache-aware samplers shrink runtime for public DLMs~\citep{wu2025fast}, and proprietary offerings such as Mercury~\citep{khanna2025mercury}, Gemini Diffusion~\citep{deepmindGeminiDiffusion}, and Seed Diffusion~\citep{song2025seed} report aggressive generation latencies. Collectively, these efforts establish diffusion-based coding as a credible alternative to purely autoregressive solutions.

\section{Preliminaries}

Denoising Diffusion Probabilistic Models (DDPMs, diffusion models)~\citep{ho2020denoising,song2020denoising} are a family of latent variable models that describe data generation as the reverse of a forward diffusion process. Given the data distribution $q(\vx_0)$, the forward diffusion process is a Markov chain that gradually injects noise into data:
\begin{equation}
    \label{eqn:ddpm_forward}
    q(\vx_{T}|\vx_0) = \prod^T_{t=1} q(\vx_t\vert\vx_{t-1}),
\end{equation}
where $q(\vx_t\vert\vx_{t-1})$ is a noise distribution. The model $p_\theta$ then learns the reverse denoising process $p_\theta(\vx_{t-1}|\vx_t)$ by aligning the marginal distributions in the reverse process to the marginal distribution in the corresponding forward step.

\paragraph{Continuous Diffusion Models.} For continuous data (e.g., images), $q(\vx_t\vert\vx_{t-1})$ is typically a Gaussian distribution $\mathcal{N}(\vx_t;\sqrt{1-\beta_t}\vx_{t-1}, \beta_t\mI)$, and the reverse process is modeled by $p_\theta(\vx_{t-1}|\vx_t) := \mathcal{N}(\vx_{t-1};\vmu_\theta(\vx_t,t), \mSigma_\theta(\vx_t,t))$.

\paragraph{Discrete Diffusion Models.} For discrete data (e.g., text tokens), suppose $\vx_t \in \{0,1\}^{V}$ is a one-hot vector drawn from a vocabulary of size $V$. In this case $q(\vx_{t}|\vx_{t-1})$ is a categorical distribution defined by a transition matrix $\mQ_t$ whose $(i,j)$-entry $[\mQ_t]_{ij}$ denotes the probability of token $i$ transitioning to token $j$. Following \citet{gong2024scaling} and \citet{shi2024simplified}, we define the transition such that with probability $\beta_t$ a token is replaced by a special \texttt{mask} token; otherwise, it stays unchanged:
\begin{equation}
    \label{eqn:discrete_forward}
    q(\vx_t|\vx_{t-1}) = Cat(\vx_t; \mQ_t^\top \vx_{t-1}), \quad \mQ_t = (1-\beta_t)\mI + \beta_t\mathbf{1}\ve_{\texttt{mask}}^\top,
\end{equation}
where $\mathbf{1}$ is an all-one vector and $\ve_{\texttt{mask}}$ is a one-hot vector where the index of \texttt{mask} token is 1. Let $\alpha_T := \prod^T_{t=1}(1-\beta_t)$ and $\bar \mQ_T:= \prod^T_{t=1} \mQ_t = \alpha_T \mI + (1-\alpha_T)\mathbf{1}\ve_{\texttt{mask}}^\top$. Then it follows that
\begin{equation}
    \label{eqn:discrete_forward_joint}
    q(\vx_T|\vx_0) = Cat(\vx_T;\bar \mQ_T^\top\vx_0) = \alpha_T\vx_0 + (1-\alpha_T)\ve_{\texttt{mask}}.
\end{equation}
Thus, at timestep $T$, with probability $(1-\alpha_T)$ the input $\vx_0$ enters the absorbing state associated with the \texttt{mask} token. This discrete-time Markov chain generalizes to a continuous Markov process. For every $0\leq s<t\leq 1$:
\begin{equation}
    \label{eqn:continuous_forward}
    q(\vx_t|\vx_s) = Cat(\vx_t;\bar \mQ(s,t)^\top \vx_s) = \frac{\alpha_t}{\alpha_s} \mI + (1-\frac{\alpha_t}{\alpha_s})\mathbf{1}\ve_{\texttt{mask}}^\top.
\end{equation}
The corresponding reverse process is given by:
\begin{equation}
    \label{eqn:continuous_backward}
    q(\vx_s|\vx_t,\vx_0) = Cat(\vx_s; \bar \mR^{\vx_0}(t,s)^\top \vx_t), \quad  \bar \mR^{\vx_0}(t,s) = I + \frac{\alpha_s - \alpha_t}{1-\alpha_t} \ve_{\texttt{mask}}(\vx_0 - \ve_{\texttt{mask}})^\top.
\end{equation}
We then use a model $f_\theta$ (typically a transformer) to approximate $f_\theta(\vx_t)\approx\vx_0$ and compute $p_\theta(\vx_s|\vx_t, f_\theta(\vx_t)) \approx q(\vx_s|\vx_t,\vx_0)$.

\paragraph{Loss Function} Following previous works~\citep{shi2024simplified,gong2024scaling}, we pick $\alpha_t:=1-t$ with $t\in(0,1)$. For a sequence of $N$ tokens $\vx:= \vx_0 = [\vx^{(1)},\ldots,\vx^{(n)}]$, the loss function is given by:
\begin{equation}
    \label{eqn:loss}
    \mathcal{L}(f_\theta, \vx_0) = \E_{t\sim U[0,1]} -\frac{1}{t} \E_{q(\vx_t|\vx_0)} \sum^N_{n=1} 1_{\vx^n_{t} = \ve_{\texttt{mask}}}(\vx^{(n)}_0)^\top(\log f_\theta(\vx_t))^{(n)},
\end{equation}
where $(n)$ indicates the $n$-th element in the corresponding vector. \Cref{eqn:loss} implies that the forward process effectively performs three steps: (1) sample a noise level $t$; (2) mask the input $\vx_0$ to obtain $\vx_t$; and (3) compute the cross-entropy loss of $f_\theta$ only on the masked tokens. When $t$ is fixed, the loss reduces to the masked-language-model objective used to train BERT~\citep{devlin2019bert}.

\section{Training}

\subsection{Overview}
CoDA builds on the Qwen3-1.7B model. We divide training into three stages: pre-training, mid-training, and post-training. The first two stages run on TPUs with a custom trainer for optimized performance, and the final post-training stage runs on GPUs using the DiffuLLaMA framework~\citep{gong2024scaling}.

\begin{figure}[htbp]
    \centering
    \includegraphics[width=0.8\linewidth]{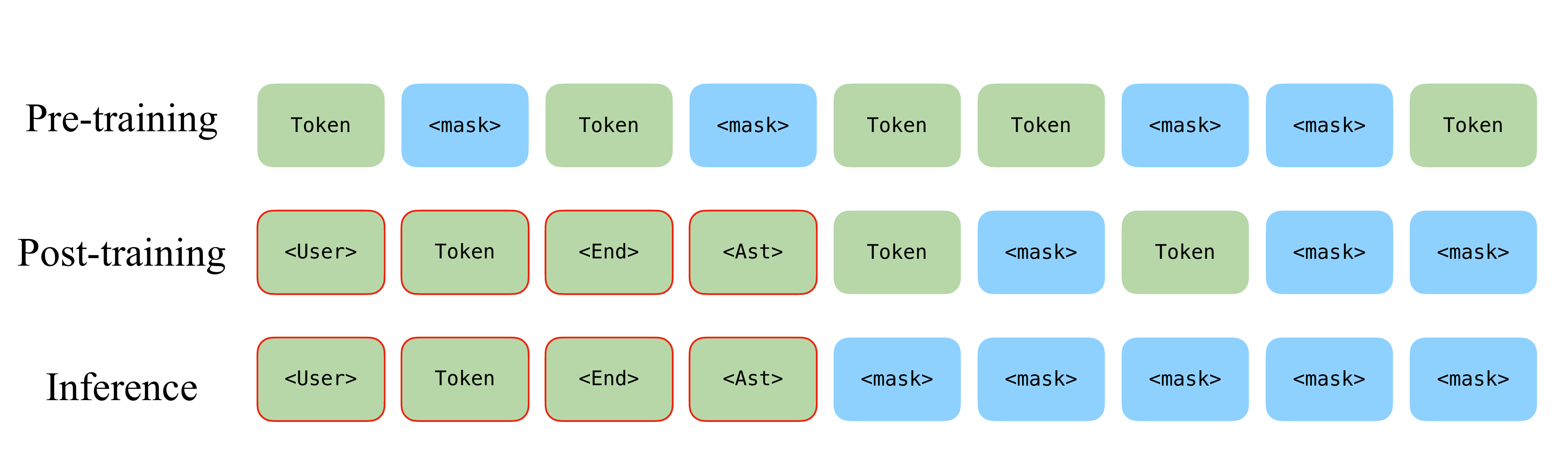}
    \caption{The masking distribution during different stages. Green tiles represent text tokens, blue tiles are mask tokens, and tiles with red border lines indicate tokens that are conditioned not to be masked. During pre-training or mid-training, masking is random. In the post-training stage, a structured masking strategy is applied. For inference, the model is conditioned on a prefix to perform infilling.}
    \label{fig:masking_distribution}
\end{figure}

A key challenge in training a DLM is the discrepancy in noise distribution between the training and inference phases. As illustrated in Figure~\ref{fig:masking_distribution}, token masking is random during pre-training and mid-training. In post-training, masking is applied randomly to the assistant's response, conditioned on a user query. During inference, however, the entire response is initialized as a contiguous block of \texttt{mask} tokens, and the model learns to denoise it conditioned on the user's query. To bridge the gap between these varying noise distributions and ensure a smooth transition across stages, we introduce a progressive masking schedule.

\paragraph{Progressive Masking Schedule}
We adopt a progressive masking schedule that gradually increases the difficulty of the training objective across epochs. In addition to standard random masking, we introduce three structured masking strategies designed to better capture prompt alignment, variable-length completion, and infilling capabilities:
\begin{itemize}
\setlength{\itemindent}{0pt}
\setlength{\leftskip}{0pt}
  \item \textbf{S1 Unmaskable Prefix}: A randomly selected prefix is left unmasked, forcing the model to consistently condition on the initial segment of the sequence. This strategy encourages stronger reliance on prompts and improves prefix-based generation.
  \item \textbf{S2 Truncated Suffix}: A randomly chosen suffix is truncated and replaced with unmaskable \texttt{pad} tokens. This setting exposes the model to incomplete contexts and prevents overfitting to full-length sequences, thereby enhancing robustness to limited or noisy inputs.
  \item \textbf{S3 Block Masking}: Rather than masking isolated tokens, we mask contiguous spans of length $k$ (with $k \in \{2, 4, 8\}$ for example) while keeping the sequence-level masking probability at $1-\alpha_t$. This block-level corruption better simulates real-world scenarios such as text infilling, style rewriting, and document editing. 
\end{itemize}

\begin{figure}[h]
    \centering
    \includegraphics[width=0.8\linewidth]{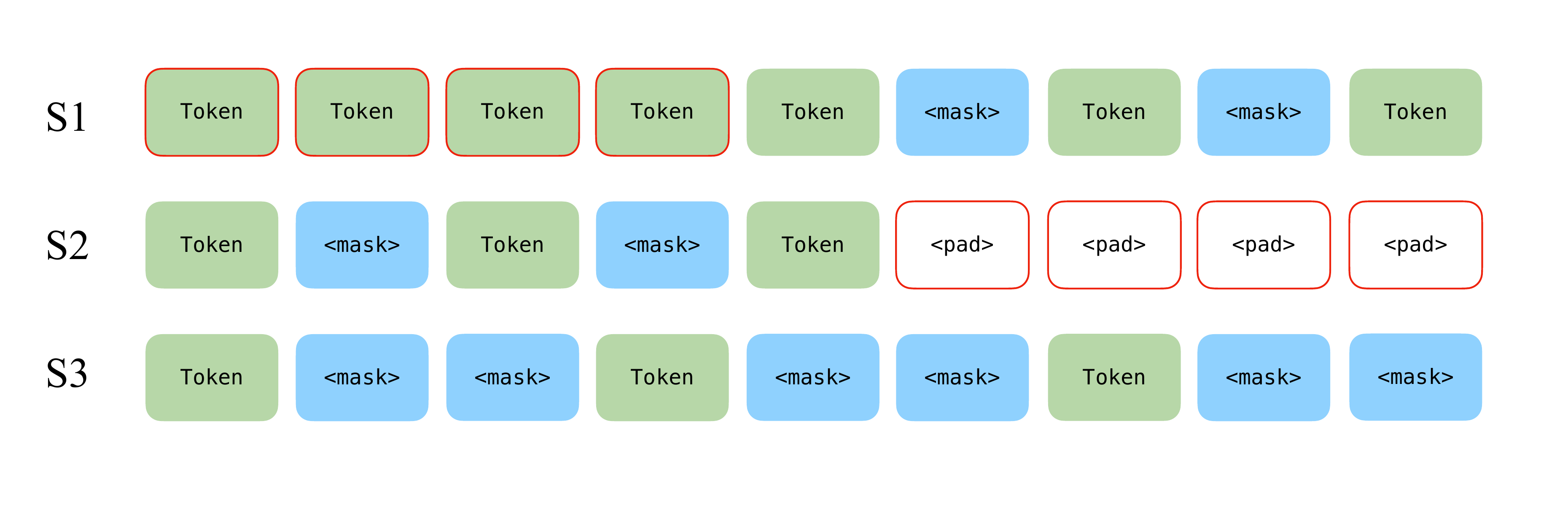}
    \caption{A visualization of the masking schedule. S1: a randomly chosen prefix is conditioned and unmaskable; S2: a randomly chosen suffix is replaced with the \texttt{pad} token and made unmaskable; S3: A block masking of size $k=2$.}
    \label{fig:masking_strategy}
\end{figure}
This progressive masking schedule both gradually increases task difficulty to move from pre-training to fine-tuning and, aligns training with downstream practical applications. The unmaskable prefix strategy (S1) encourages the model to rely on prompts, stabilizing optimization and improving prefix-conditioned generation during supervised fine-tuning. The truncated suffix strategy (S2) teaches the model to handle sequences of varying length and prevents overfitting to full-length examples, which benefits tasks involving truncated queries or constrained context windows. The block masking strategy (S3) exposes the model to contiguous spans of missing tokens, closely matching infilling, editing, and style-transfer scenarios. Together these strategies enhance robustness to prompts, adaptability to variable-length inputs, and performance on editing or completion tasks. Increasing the masking difficulty across epochs yields a smooth progression from easier to harder training conditions.

\subsection{Infrastructure}
Our training infrastructure runs on Google's TPU VMs. We enable PyTorch-based training with the PyTorch/XLA library and build on \texttt{torchprime}~\citep{torchprime} to develop a custom trainer tailored to our workflow. The trainer provides robust distributed checkpointing for efficient saving and loading of model states across TPU cores, integrates the progressive masking schedule directly into the training loop, and drives high-throughput data pipelines capable of processing up to one billion tokens per update.

\subsection{Pre-training}
During pre-training, transitioning a causal LM to a DLM requires exposing the model to massive data so it shifts from causal attention and autoregressive decoding to bidirectional attention and unmasking. We follow the adaptation strategy of \citet{gong2024scaling} but omit attention-mask annealing because FlashAttention does not support customized masking.

\subsubsection{Pre-training Data Corpora}
Our pre-training corpus is a comprehensive collection of text and code totaling 179.27 billion tokens, designed to provide broad knowledge and strong reasoning capabilities. After data cleaning, deduplication, and pre-tokenization, the corpus composition appears in \Cref{tab:pre_train_data}.

A significant portion of the data comes from web text and source code.
For web data, we use \texttt{dclm-baseline-1.0}~\citep{li2024datacomp}, a large-scale web crawl that contributes 60.17 billion tokens. 
For source code, we include two subsets from The Stack v2 dataset~\citep{lozhkov2024starcoder}: 
1. \texttt{the-stack-v2 (Python)} with 50.75 billion tokens of Python code, 
2. \texttt{the-stack-v2-train-smol} with 38.22 billion tokens spanning multiple programming languages.

To bolster mathematical and scientific reasoning, we include several curated datasets.
OpenWebMath~\citep{paster2023openwebmath} offers 12.98 billion tokens of mathematical text sourced from the web. Additionally, we use the arXiv subset of RedPajama~\citep{weber2024redpajama}, which provides 9.18 billion tokens from academic papers available on arXiv.
For mathematical problem solving, we incorporate the DeepMind Mathematics dataset~\citep{Saxton2019AnalysingMR}, contributing 3.24 billion tokens.
To ensure a solid foundation of general knowledge, we add the English Wikipedia dump~\citep{wikidump}, which supplies 5.41 billion tokens of high-quality encyclopedic text.

\begin{table}[hbt!]
    \centering
    \caption{Pre-training data corpus with categories and token counts (in billions)}
    \label{tab:pre_train_data}
    \begin{tabular}{lccc}
        \toprule
        \textbf{Dataset} & \textbf{Category} & \textbf{Token (Billion)} \\ 
        \midrule
        dclm-baseline-1.0 & Text (Web) & 60.17 \\ 
        The Stack v2 (Python) & Code & 50.75 \\
        The Stack v2 (train-smol) & Code & 38.22 \\ 
        OpenWebMath & Text (Mathematics) & 12.98 \\
        Redpajama Arxiv & Text (Academic) & 9.18 \\ 
        Wikipedia (English) & Text (Encyclopedia) & 5.41 \\
        DeepMind Mathematics & Text (Mathematics) & 3.24 \\
        \bottomrule
    \end{tabular}
\end{table}

\subsubsection{Pre-training Recipe}
We run pre-training on a TPU v4-1024 VM with Torch XLA and FSDP. The global batch size is 512 with a sequence length of 8192, totaling around four million tokens per update. We use the Adafactor optimizer~\citep{shazeer2018adafactor} with linear learning-rate scheduling and set the peak learning rate to $3\times10^{-4}$. With probability $0.01$, each batch applies masking strategy S1 or S2; independently, with the same probability, S3 applies to the batch. Training spans 50{,}000 steps, just over one epoch.

\subsection{Mid-training}
To strengthen the model’s capacity to handle complex masking patterns and to bridge pre-training with downstream fine-tuning, we extend training with high-quality corpora comprising both coding and textual datasets. 

\subsubsection{Mid-training Data Corpora}
During mid-training, we incorporate two high-quality textual corpora totaling 12.42 billion tokens: RedPajama Arxiv~\citep{weber2024redpajama} and Gutenberg~\citep{stroube2003literary}. We also include three coding corpora amounting to 8.55 billion tokens: the \texttt{python-edu} subset of SmolLM-Corpus~\citep{benallal2024smollmcorpus}, the OpenCoder Annealing Corpus~\citep{Huang2024OpenCoderTO}, and Python code extracted from The Stack v2~\citep{lozhkov2024starcoder}. We reuse a small slice of The Stack v2 data from pre-training to ensure a smooth transition between stages. \Cref{tab:mid_train_data} provides a detailed breakdown of the mid-training datasets.
\begin{table}[hbt!]
\centering
\caption{Mid-training data corpora with categories and token counts (in billions).}
\label{tab:mid_train_data}
\begin{tabular}{lcc}
\toprule
\textbf{Dataset} & \textbf{Category} & \textbf{Tokens (B)} \\
\midrule
Redpajama Arxiv                & Text (Academic) & 9.18 \\
Gutenberg                      & Text (ebooks)       & 3.24 \\
SmolLM-Corpus (python-edu)      & Code                & 1.88 \\
OpenCoder (opc-annealing-corpus) & Code              & 6.47 \\
The Stack v2 (Python)          & Code                & 0.20 \\
\bottomrule
\end{tabular}
\end{table}
\subsubsection{Mid-training Recipe}
We run mid-training on a TPU v4-512 with Torch XLA and FSDP. The global batch size is 256 with a sequence length of 8192, yielding around two million tokens per step. We use the Adafactor optimizer with linear learning-rate scheduling and set the peak learning rate to $2\times 10^{-4}$. Training spans 50{,}000 steps (roughly five epochs). Throughout mid-training we evenly shuffle the datasets, maintain five full passes, and gradually increase the probabilities of applying S1, S2, and S3 according to a fixed curriculum (see \cref{tab:mid_train_mask}).

\begin{table}[hbt!]
\centering
\caption{Progressive masking schedule over mid-training epochs. Probabilities indicate the proportion of sequences affected by each strategy (S1, S2, S3).}
\label{tab:mid_train_mask}
\begin{tabular}{c c}
\toprule
\textbf{Epoch} & \textbf{S1 : S2 : S3} \\
\midrule
1 & $1\% \; : \; 1\% \; : \; 5\%$ \\
2 & $5\% \; : \; 5\% \; : \; 10\%$ \\
3 & $10\% \; : \; 10\% \; : \; 15\%$ \\
4 & $15\% \; : \; 15\% \; : \; 20\%$ \\
5 & $20\% \; : \; 20\% \; : \; 25\%$ \\
\bottomrule
\end{tabular}
\end{table}

\subsection{Post-training}
After mid-training, we conduct supervised fine-tuning as the post-training stage so the model can address real-world coding tasks. Post-training runs on a single H200 cluster. We use the stage 1 and stage 2 SFT subsets of the OpenCoder dataset~\citep{Huang2024OpenCoderTO} and truncate sequences to 1024 tokens. For stage 1, we train for 10 epochs (with early stopping) using the AdamW optimizer and a cosine scheduler, using a peak learning rate of $2\times 10^{-5}$ with a $10\%$ warm-up. During the first epoch, we gradually transition from unconditional training to conditioning on the user prompt by linearly increasing the conditioned span. The global batch size is 176, and we reuse the same hyperparameters for stage 2 SFT.

\section{Experiments}

\subsection{Model Evaluation}
We evaluate CoDA against other DLMs and similarly sized AR models on two coding benchmarks: Humaneval~\citep{chen2021codex} and MBPP~\citep{austin2021program}. For both datasets, we also report the EvalPlus-enhanced variants provided by \citet{eval-harness}. All evaluations use pass@1 as the primary metric. Our evaluation harness builds on Dream~\citep{ye2025dream} and Qwen2.5-Coder~\citep{hui2024qwen2}, with adaptations to support diffusion decoding and data parallelism.

For CoDA, we cap the maximum number of generated tokens at $768$ and align the diffusion schedule to the sequence length, i.e., a single token is produced at each diffusion step. We adopt the confidence-based sampling strategy described in \citet{ye2025dream}, which reweights denoising updates using per-token posterior entropy. Self-reported AR baselines use the default configuration provided by their released inference environments, including nucleus sampling and temperature values recommended by their maintainers.

\begin{table}[htbp]
    \centering
    \caption{Comparison of code generation performance on Humaneval and MBPP. Evalplus scores are calculated as an average of pass@1 scores on plus-enhanced variants. Bold numbers indicate metrics where CoDA models achieve the strongest diffusion-model result. $\ast$ represents self-reported scores.}
    \label{tab:code_generation_benchmark}
    {
    \setlength{\tabcolsep}{8pt}
    \begin{tabular}{@{}lccccc@{}}
        \toprule
        \multirow{2}{*}{\textbf{Model}} & \multicolumn{2}{c}{\textbf{Humaneval}} & \multicolumn{2}{c}{\textbf{MBPP}} & \multirow{2}{*}{\textbf{Evalplus}} \\
        \cmidrule(lr){2-3} \cmidrule(lr){4-5}
        & \textbf{-} & \textbf{Plus} & \textbf{-} & \textbf{Plus} & \\
        \midrule
        \multicolumn{6}{l}{\textit{Diffusion Models}} \\
        \midrule
        CoDA-1.7B-Base & 29.3 & 23.8 & 35.2 & 46.0 & 34.9 \\
        CoDA-1.7B-Instruct & 54.3 & 47.6 & 47.2 & \textbf{63.2} & \textbf{55.4} \\
        Dream-7B-Base & 56.7 & 50.0 & 68.7 & 57.4 & 53.7 \\
        Dream-7B-Instruct & 57.9 & 53.7 & 68.3 & 56.1 & 54.9 \\
        LLaDA-8B-Instruct & 35.4 & 31.7 & 31.5 & 28.6 & 30.2 \\
        \midrule
        \multicolumn{6}{l}{\textit{AR Models of similar size}} \\
        \midrule
        Qwen3-1.7B* & 66.5 & 61.6 & 46.2 & 65.9 & 63.8 \\
        Qwen2.5-Coder-1.5B & 43.9 & 36.6 & 69.2 & 58.6 & 47.6 \\
        Qwen2.5-Coder-1.5B-Instruct & 70.7 & 66.5 & 69.2 & 59.4 & 62.3 \\
        Gemma-3-1B-it* & 39.6 & 35.4 & 39.4 & 63.5 & 49.5 \\
        LLaMA-3.2-1B-Instruct* & 35.4 & 31.1 & 24.4 & 53.7 & 42.4 \\
        \bottomrule
    \end{tabular}}
\end{table}

\Cref{tab:code_generation_benchmark} summarizes the performance of all systems on the evaluation suite. Across diffusion models, CoDA-1.7B-Instruct narrows much of the gap to larger Dream models while outperforming other DLMs on most reported metrics. The $25$-point improvement of CoDA-1.7B-Instruct over CoDA-1.7B-Base on Humaneval (pass@1) illustrates the potential of instruction tuning. Although Dream-7B-Instruct remains the strongest diffusion baseline on MBPP-Instruct, CoDA-1.7B-Instruct delivers comparable EvalPlus scores within a significantly smaller parameter count. When compared with AR models of similar scale, CoDA-1.7B-Instruct trails the Qwen model series but surpasses the Gemma or LLaMA models. These results highlight that diffusion decoding can remain competitive at small model sizes.

\subsection{Inference Scaling Dynamics}
We investigate how the number of diffusion steps affects the inference time of CoDA-1.7B-Instruct and evaluate the model's performance across diffusion steps ranging from 32 to 1024. We conducted experiments on a single NVIDIA A100 40GB. We applied a KV cache variant to accelerate inference and adopted confidence-aware parallel decoding to balance the inference efficiency and quality following~\cite{wu2025fast}. The model inferred on the Humaneval dataset with a 768-token budget. We run each sample for two trials when measuring the inference time to reduce variance.

\begin{figure}[htbp]
    \centering
    \includegraphics[width=0.6\textwidth]{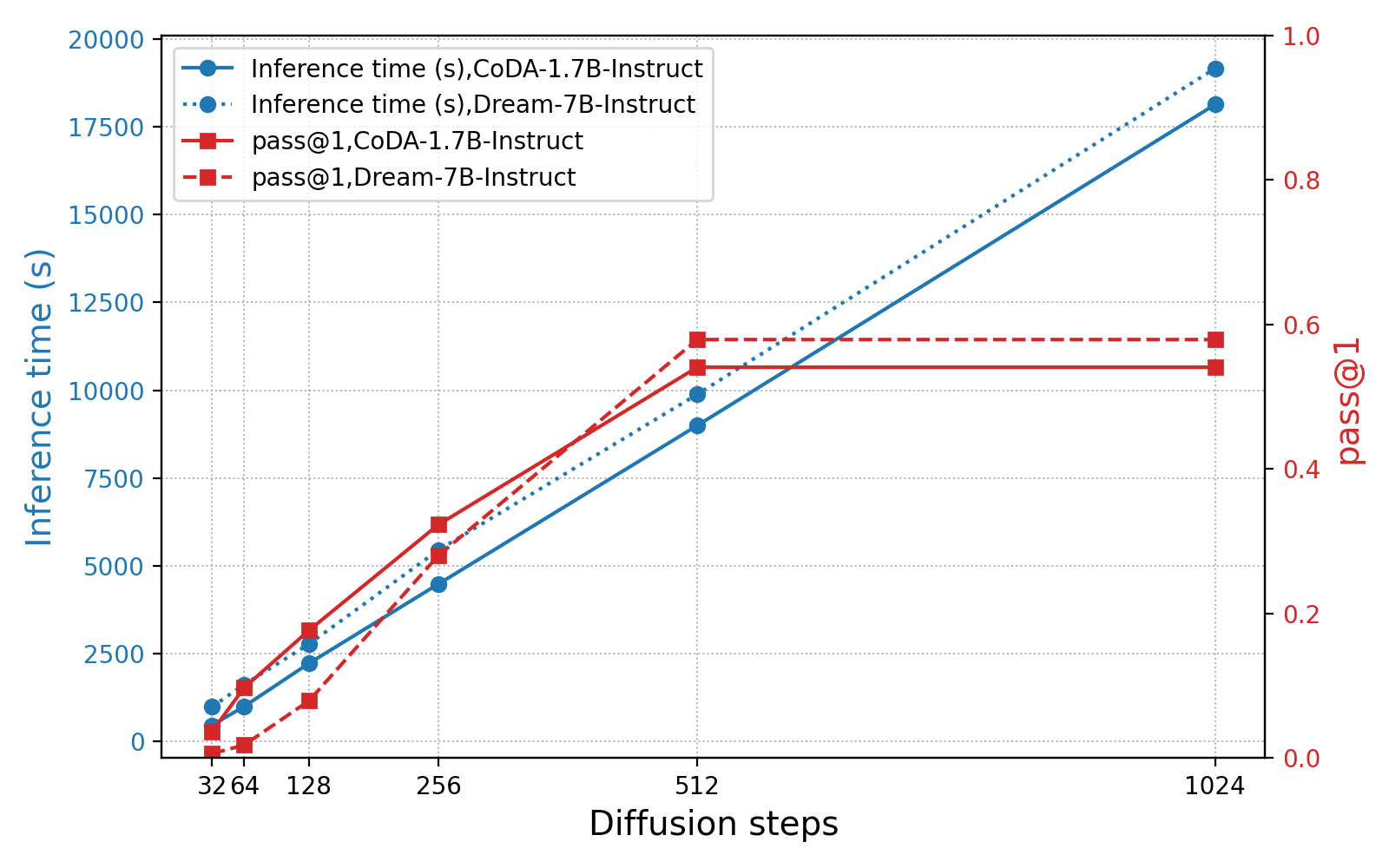}
    \caption{Relationship between diffusion steps, inference time, and CoDA-1.7B-Instruct performance. Inference time is measured by the total inference time on the Humaneval dataset. Model performance is measured by pass@1 on the same dataset with a 768-token budget.}
    \label{fig:speed_accuracy}
\end{figure}
Figure~\ref{fig:speed_accuracy} illustrates the trade-off between inference time and model performance for CoDA-1.7B-Instruct across varying diffusion steps. As shown by the blue curve, inference time for both CoDA-1.7B-Instruct and Dream-7b-Instruct increases almost linearly with diffusion steps, highlighting the computational cost of lengthening the diffusion trajectory. CoDA-1.7B-Instruct achieves $39.64\%$ lower latency than Dream-7B-Instruct across the same range of diffusion steps, due to its smaller parameter footprint. The red curves track task performance (pass@1 on Humaneval with a 768-token budget). Both CoDA-1.7B-Instruct and Dream-7B-Instruct show steady improvements as steps increase, but gains saturate around $512$ steps. Notably, CoDA-1.7B-Instruct surpasses Dream-7B-Instruct at smaller step counts despite its more compact size, we attribute this behavior to our block masking strategies, since when the diffusion step is much smaller than the total tokens to generate, we are essentially unmasking multiple tokens at a time.

\subsection{Discussion}
\Cref{tab:code_generation_benchmark} shows that CoDA-1.7B-Instruct is competitive with other diffusion models of larger size while staying within a 1.7B-parameter budget. Instruction tuning paired with confidence-guided sampling closes most of the gap to Dream-7B-Instruct on Humaneval and EvalPlus, validating that diffusion adaptation can unlock strong coding accuracy without resorting to heavyweight backbones. Against autoregressive (AR) coders, CoDA retains a balanced profile: it trails the strongest AR baselines on Humaneval but delivers competitive MBPP performance, suggesting that diffusion decoding provides complementary inductive biases even when model scale is modest.

The diffusion schedule analysis in \Cref{fig:speed_accuracy} highlights a second advantage of compact DLMs. Because the inference latency scales roughly linearly with the number of steps, yet accuracy saturates beyond 512 steps. This observation indicates that practical deployments can adopt adaptive schedules that cap the number of steps when confidence is high, yielding faster responses than larger autoregressive models that require long decoding traces. Our open-source TPU pipeline further demonstrates that diffusion pre-training and instruction tuning remain tractable when optimized end-to-end on modern pods, enabling rapid iteration on architectural variations and sampling strategies.

\section{Conclusion}
We present CoDA, a fast 1.7B diffusion language model for code that demonstrates competitive performance with several 7B models while maintaining efficiency suitable for lightweight hardware budgets. By releasing model checkpoints, training pipelines, and evaluation harnesses, we aim to ease the challenge for the community to explore diffusion-based coding assistants and accelerate progress in this emerging paradigm. Looking forward, we plan to explore hybrid diffusion/AR decoding, reinforcement learning–based fine-tuning, and more refined pre-training strategies to push the frontier of diffusion coders toward higher accuracy, faster inference, and broader applicability.

\section*{Acknowledgement}
We would like to thank Lingpeng Kong for insightful discussions and Jialei Chen for technical support in TPU.
\bibliographystyle{abbrvnat}
\bibliography{reference}

\end{document}